%% file: meaning_identity.tex
\pdfoutput=1
\documentclass[11pt]{article}
\usepackage[margin=1in]{geometry}
\usepackage[T1]{fontenc}
\usepackage[utf8]{inputenc}
\usepackage{amsmath}
\usepackage{amssymb}
\usepackage{bm}
\usepackage{newtxtext,newtxmath}
\usepackage{graphicx}
\usepackage{booktabs}
\usepackage{microtype}
\usepackage{xcolor}
\usepackage{url}
\usepackage{natbib}
\usepackage{caption}
\usepackage{titlesec}
\titleformat*{\section}{\large\bfseries}
\titleformat*{\subsection}{\normalsize\bfseries}
\usepackage[hidelinks]{hyperref}
\input{math_commands}

\title{\bfseries Computation Over Geometry:\\Meaning Identity Is Computed,\\Not Shipped in the Embeddings}
\author{Jiaqi Deng\\ \small Independent Researcher\\ \small \texttt{djq627@163.com}}
\date{}

\begin{document}
\maketitle

\begin{abstract}
Meaning identity---whether two sentences say the same thing after the wording has changed---is treated throughout retrieval and {RAG} as a geometric fact about independently encoded sentence vectors.
We show that, for frozen off-the-shelf encoders and language models, it is not: identity is \emph{computed} when both sentences occupy the same forward pass, and it is not a property of the embedding geometry those systems ship.
On overlap-matched {PAWS-X}, purpose-built encoders ({BGE}, {E5}, {GTE}, {MiniLM}, and {E5}-Mistral-7B) reach English confirm {AUC} only $0.55$--$0.65$ (dense peak $0.70$; $n{=}900$).
Independently encoded last-token states of Llama~3, Mistral, and {Qwen} do no better; a linear probe on the concatenation of the two vectors (\emph{late fusion}) remains near chance.
The same probe on the last token of a \emph{joint} forward pass over both sentences reaches $0.90$--$0.96$ from 1.5{B} through 32{B}, and shuffling sentence~B collapses it to chance; the signal is mid-depth, saturates near $0.94$ by 3{B}, and is already present, more weakly, in {GPT-2}~{XL} ($0.76$).
The pattern is not a Llama-family artefact: the same four readouts on non-{Qwen}/{Llama}/{Mistral} causal models, on bidirectional encoders ({DeBERTa}, {RoBERTa}), and on encoder--decoders ({Flan}-{T5}, {T5}, {BART}) again give joint $\gg$ late fusion with shuffle near chance.
No fixed or linear reader over frozen independent encodings---cosine, token-level MaxSim, late fusion---makes identity accessible at any training size; nonlinear pair readers trained from scratch, an {MLP} over the two vectors or a cross-attention module over the frozen token sequences, recover part of it only on the full 49k-pair {PAWS} train split ($0.68$--$0.87$), forty times the pairs the joint probe needs.
Off-the-shelf rerankers split the same way: {BGE}-reranker-large reaches $0.94$, but {MS-MARCO} and Jina rerankers stay in the dense band ($0.55$--$0.64$): joint scoring is necessary, not sufficient.
Independently trained families compute the \emph{same} relation: their joint scores agree, their errors co-occur, and a 1.5{B} joint reader distilled from unlabelled teacher-scored pairs recovers it, while no linear function of the teacher's own independent vectors can.
A bi-encoder can be fine-tuned to fit {PAWS} ({AUC} $0.87$--$0.93$), but the fit is a {PAWS}-specific criterion: transfer to overlap-matched {QQP} drops and {STS-B} Spearman falls by about $0.25$.
Off-the-shelf cosine compares wording neighbourhoods; identity is a cheap computed operator, not a property of either sentence's vector.
\end{abstract}

\section{Introduction}

A sentence embedding is an invitation to treat meaning as a point.
Once each sentence has a vector, ``same meaning'' becomes cosine, and the industrial stack---dense retrieval, duplicate-question detection, {RAG}---is built on that geometry~\citep{reimers2019sentence,muennighoff2023mteb,gao2024rag}.
The invitation is so familiar that it is easy to miss the empirical claim it encodes: that identity of meaning is a property of a single sentence's representation, recoverable by comparing two such properties.

That claim can be false even if embeddings are useful.
They can rank topical neighbours, cluster documents, and win {STS}~\citep{cer2017semeval} while still being nearly blind to the distinction {PAWS} was built to isolate: two sentences that share almost all of their words, one of which is a paraphrase and one of which is not~\citep{zhang2019paws,yang2019pawsx}.
If identity is not in the vectors, then every system that first encodes and then compares is looking in the wrong place for a large class of hard pairs.

We separate two pictures, on identical items, with a metric whose chance level is exactly $0.5$:

\paragraph{Geometry.}
Each sentence carries an identity. Cosine, a pair probe, or late fusion of the two vectors should recover it.

\paragraph{Computation.}
Identity is a relation the network computes when both sentences occupy the same context.
It is then readable from the last token of a concatenated forward pass, and \emph{not} from any fixed or linear function of the two separately encoded states.

The pictures are not rhetorical: late fusion gives a linear probe both vectors and no cross-attention, so if geometry were merely ``misaligned with cosine'' it would close the gap; a shuffled-partner joint pass preserves format, language, and the marginal of sentence~B while destroying pairing, so a probe reading a template rather than the pairing would survive it.

\begin{table}[t]
\caption{English confirm {AUC} at the layer of peak joint probe. Late fusion and shuffle are the two falsifiers of a computational reading. Chance $=0.5$. {Qwen}2.5-32{B} joint 95\% CI is $[0.897,0.969]$; 14{B} is $[0.925,0.971]$.}
\label{tab:joint}
\begin{center}
\scriptsize
\setlength{\tabcolsep}{3.2pt}
\begin{tabular}{llccccc}
\toprule
Model & layer & Cos & Sep & Late & \textbf{Joint} & Shuf \\
\midrule
{Qwen}2.5-14{B} & L36 & $.614$ & $.615$ & $.541$ & $\mathbf{.950}$ & $.496$ \\
{Qwen}2.5-7{B}-Inst & L21 & $.571$ & $.612$ & $.529$ & $\mathbf{.961}$ & $.526$ \\
{Qwen}2.5-1.5{B}-Inst & L21 & $.565$ & $.609$ & $.542$ & $\mathbf{.947}$ & $.509$ \\
{Qwen}2.5-7{B} & L21 & $.556$ & $.572$ & $.483$ & $\mathbf{.941}$ & $.520$ \\
{Qwen}2.5-32{B} & L58 & $.569$ & $.557$ & $.493$ & $\mathbf{.938}$ & $.473$ \\
{Qwen}2.5-3{B} & L27 & $.526$ & $.538$ & $.555$ & $\mathbf{.937}$ & $.521$ \\
{Qwen}2.5-1.5{B} & L21 & $.560$ & $.549$ & $.529$ & $\mathbf{.916}$ & $.458$ \\
Mistral-7{B}-v0.3 & L16 & $.553$ & $.563$ & $.508$ & $\mathbf{.915}$ & $.499$ \\
Llama-3-8{B} & L16 & $.530$ & $.514$ & $.473$ & $\mathbf{.897}$ & $.542$ \\
Phi-2 & L24 & $.607$ & $.597$ & $.524$ & $\mathbf{.893}$ & $.520$ \\
{GPT-J}-6{B} & L14 & $.565$ & $.508$ & $.532$ & $\mathbf{.868}$ & $.538$ \\
{Qwen}2.5-0.5{B}-Inst & L18 & $.563$ & $.557$ & $.497$ & $\mathbf{.863}$ & $.494$ \\
SmolLM2-1.7{B} & L18 & $.528$ & $.616$ & $.581$ & $\mathbf{.843}$ & $.498$ \\
{OPT}-6.7{B} & L24 & $.607$ & $.585$ & $.561$ & $\mathbf{.829}$ & $.466$ \\
{GPT-Neo}-1.3{B} & L12 & $.539$ & $.582$ & $.546$ & $\mathbf{.815}$ & $.511$ \\
{Qwen}2.5-0.5{B} & L18 & $.529$ & $.544$ & $.568$ & $\mathbf{.789}$ & $.503$ \\
TinyLlama-1.1{B} & L16 & $.631$ & $.641$ & $.543$ & $\mathbf{.778}$ & $.441$ \\
Phi-1.5 & L12 & $.548$ & $.577$ & $.539$ & $\mathbf{.778}$ & $.497$ \\
{GPT-2} {XL} & L24 & $.521$ & $.566$ & $.513$ & $\mathbf{.756}$ & $.457$ \\
SmolLM2-360{M} & L24 & $.582$ & $.550$ & $.513$ & $\mathbf{.753}$ & $.530$ \\
\bottomrule
\end{tabular}
\end{center}
\end{table}

\begin{figure}[t]
\centering
\includegraphics[width=0.85\linewidth]{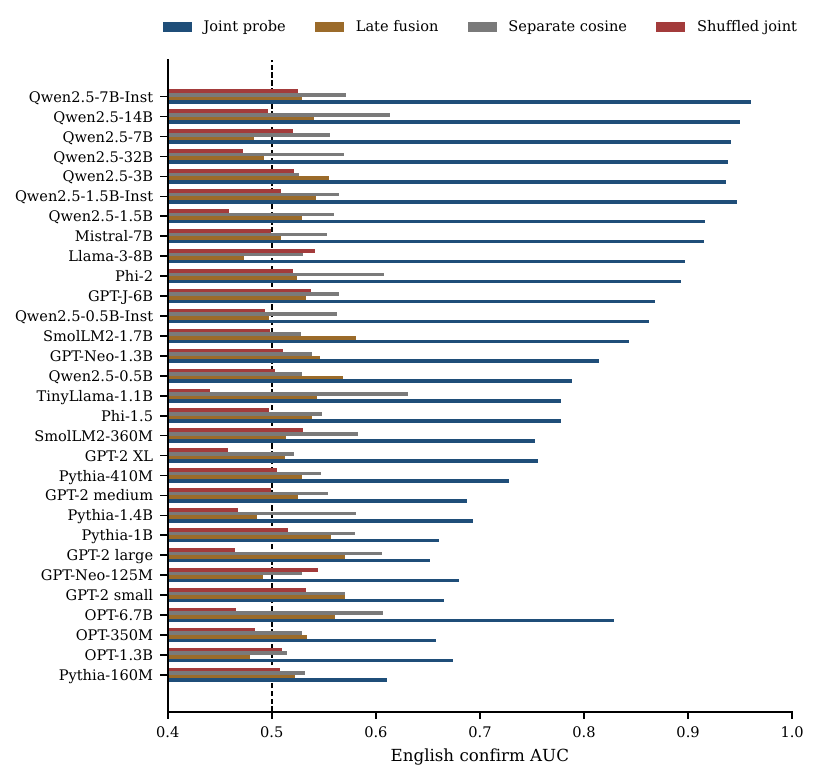}
\caption{English confirm {AUC} at each model's peak joint layer (overlap-matched {PAWS-X}).
Joint sits above late fusion, cosine, and shuffle, which stay near chance.
Same numbers as Table~\ref{tab:joint}.}
\label{fig:bars}
\end{figure}

We find computation, not geometry, across modern open families (Table~\ref{tab:joint}, Figure~\ref{fig:bars}).
On overlap-matched English {PAWS-X}, Llama~3~8{B}, Mistral~7{B}, and {Qwen}2.5 from 1.5{B} to 32{B} all yield joint confirm {AUC} $0.90$--$0.96$ at mid-depth, while separate cosine, a separate pair probe, and late fusion remain in $0.47$--$0.61$; partner shuffle returns to $\approx 0.50$; dedicated encoders never leave the $0.55$--$0.70$ band.
{GPT-2} is not a counterexample species: the same joint circuit appears more weakly, and only {XL} clearly pulls away ($0.76$).
Outside that stack the same protocol holds for Phi-2, {GPT-J}-6{B}, {OPT}-6.7{B}, {GPT-Neo}, Pythia, and {SmolLM}2 (joint $0.75$--$0.90$), for mean-pooled {DeBERTa}/{RoBERTa} encoders ({DeBERTa}-v3 $0.92$, nli-{DeBERTa} $0.94$), and for encoder--decoders ({Flan}-{T5}-base $0.95$; {T5}-small / {BART}-base $0.64$--$0.66$).
A nonce-substitution control keeps joint above late fusion with shuffle near chance on {Qwen} and on non-{Qwen} causals.
{Qwen} base models can also \emph{say} the answer (zero-shot verdict {AUC} $0.92$--$0.99$ from 3{B} to 32{B}); Llama~3 and Mistral compute the relation at $0.90$ while their verbal verdict stays $\approx 0.61$.

Two further findings sharpen the claim from ``not geometry'' to ``one computed operator''.
First, six independently trained models ({Qwen}, Llama, Mistral) compute measurably the \emph{same} relation: their item-level scores agree far beyond the labels, their errors co-occur, and the relation is symmetric and graded (\S\ref{sec:universal}).
Second, the operator is cheap and deployable: a 1.5{B} joint reader trained only on the \emph{consensus} of three cross-family teachers over unlabelled text matches the 7{B} teacher on {PAWS} ($0.959$) and repairs a cosine FAQ stack on wording twins (trapped top-1 $0.47 \to 0.88$) without touching the retrieval index (\S\ref{sec:world}).

The implication is not that embeddings should be discarded, but that they should not be asked to certify identity: a stack that retrieves ``the same question asked differently'' is asking a bi-encoder to do a cross-encoder's job, and the shipped geometry of current {LMs} does not contain that job.

\section{Related work}

Sentence embeddings trained with siamese or contrastive objectives~\citep{reimers2019sentence,gao2021simcse,wang2022e5,xiao2024cpack} dominate retrieval benchmarks~\citep{thakur2021beir,muennighoff2023mteb}.
The retrieval literature already knows that \emph{cross}-encoders, which read the pair jointly, outperform bi-encoders on reranking~\citep{nogueira2019passage,khattab2020colbert}.
That fact is usually treated as a compute/quality tradeoff.
We show it is sharper: {BGE} rerankers reach the joint-probe band on overlap-matched {PAWS}, while {MS-MARCO} and Jina rerankers do not---joint scoring alone does not imply identity.
We treat it as evidence about \emph{where the information is accessible}: if every fixed or linear reader over two frozen encodings stays near chance at any training size, while pair computations trained from scratch over the same encodings need forty times the pairs of a joint-pass probe to approach it, the gap is not a scoring detail---identity becomes accessible only through a computation over the pair (the claim is about accessibility to trained readouts, not information-theoretic absence).
Late interaction~\citep{khattab2020colbert} still encodes independently, then scores a function of the two sequences; our late-fusion probe and MaxSim are its sentence- and token-level analogues.

Probing work asks what is linearly readable from {LM} states~\citep{alain2017understanding,tenney2019bert,hewitt2019structural,pimentel2020information,belinkov2019analysis}.
The linear representation hypothesis~\citep{park2024linear} concerns directions \emph{inside} a model.
Recent probes on modern open models sharpen the same map: lexical identity is linearly strongest early and weakens with depth~\citep{li2026sleuthing}, while semantic and syntactic signals occupy partly separable mid-depth regimes~\citep{acevedo2026differential}, and lexical overlap continues to distort independent-embedding similarity under paraphrase stress tests~\citep{rizwan2026lexicality}.
Our question is narrower and prior to that geometry: whether the \emph{relation} ``these two sentences are paraphrases'' is a function of two independent points at all.

{PAWS} and {PAWS-X}~\citep{zhang2019paws,yang2019pawsx} adversarialise bag-of-words overlap.
Broader paraphrase evaluation has since moved beyond a single classification set~\citep{michail2025paraphrasus}; we stay with overlap-matched {PAWS}-style identity because that is exactly where shipped bi-encoder geometry is asked to certify sameness and fails.
We go further and \emph{match} overlap between paraphrase and non-paraphrase cells, so a scorer that only sees word overlap is at chance---the setting in which identity, if it exists as geometry, would have to show itself.

\section{Protocol}
\label{sec:protocol}

\paragraph{Data.}
We sample {PAWS-X} English and Chinese pairs into discovery / development / confirm splits, and inside each language$\times$split cell we match lexical overlap of paraphrases and non-paraphrases (English confirm overlap gap typically $<0.02$), so the overlap-only {AUC} on English confirm is $\approx 0.47$--$0.55$.
Joint-probe confirm cells contain $150$--$500$ pairs ($600$--$2000$ discovery); the 32{B} cell is the smallest ($150$), and its interval is reported wherever it is used.
We report {AUC} rather than forced-choice accuracy so that a constant anisotropic offset cannot move the number~\citep{ethayarajh2019contextual}, and bootstrap 95\% intervals on confirm; chance is exactly $0.5$ regardless of class balance.

\paragraph{Four pair readouts, one item set.}
For each frozen causal {LM} we score the same pairs four ways.
\emph{Separate cosine}: last-token states from independent forwards, cosine.
\emph{Separate probe}: a logistic probe on $(\mathbf{h}_1 \odot \mathbf{h}_2,\,|\mathbf{h}_1-\mathbf{h}_2|)$ after a {PCA} map fit on discovery.
\emph{Late fusion}: the same probe class on $[\mathbf{h}_1;\mathbf{h}_2]$---both vectors present, no mixing inside the {LM}.
\emph{Joint probe}: the two sentences are concatenated with a language-specific template that is \emph{not} a yes/no question (\texttt{A: \ldots} / \texttt{B: \ldots}); we probe the last token of that joint pass.
Probes are fit on discovery, regularisation is selected on development, and scores are reported only on confirm.
Table~\ref{tab:joint} lists each model at its layer of peak confirm joint {AUC}; locking the layer on development instead picks the same layer for 26 of 30 models and changes confirm joint {AUC} by $0.001$ on average and at most $0.011$ (32{B}: L48, $0.927$ instead of L58, $0.938$).

\paragraph{Shuffle control.}
Sentence~B is replaced by another sentence from the same language and split: format, language, and the marginal of~B are preserved; pairing is destroyed.
A probe that reads ``two sentences appeared'' rather than ``these two belong together'' will survive shuffle.

\paragraph{Depth.}
We record last-token states at a fixed grid of relative depths, plus the final-norm input (prenorm, captured by hook---{HuggingFace} \texttt{hidden\_states[-1]} is already post-norm) and output (postnorm).

\paragraph{Generative channel, separately.}
A zero-shot yes/no verdict, length-normalised {PMI}, and conditional surprisal are scored on a smaller overlap-matched draw; unlike the joint probe, these ask the model to \emph{speak}.

\paragraph{Encoders and rerankers.}
Nineteen public dense checkpoints ({BGE}, {E5}, {GTE}, MiniLM, mpnet, multilingual {E5}, {E5}-Mistral-7B) and six cross-encoders (three {BGE} rerankers, two {MS-MARCO} MiniLM rerankers, and Jina reranker-v2) are scored on a dedicated overlap-matched bank with $n{=}900$ confirm pairs per language, fixed sampling seeds, and English overlap matching (confirm overlap gap $-0.002$); bi-encoders and rerankers score the same items.

\paragraph{Pre-registered readings.}
\textbf{(P1)} Computation: joint $\gg$ late fusion, and late fusion stays near the separate-probe band.
\textbf{(P2)} Geometry: late fusion $\approx$ joint, both well above chance.
\textbf{(P3)} Shuffle must fall to chance.
\textbf{(P4)} Report the first depth whose confirm 95\% interval excludes $0.55$.

\section{Geometry does not carry identity}
\label{sec:geo}

Dedicated encoders fail on overlap-matched confirm.
The strongest English cosine is $0.701$ ({BGE}-large-zh); the strongest English-\emph{trained} checkpoint is mpnet-base ($0.653$).
Typical retrieval encoders sit at $0.55$--$0.65$; scaling to {E5}-Mistral-7B does not break the band ($0.613$).
A linear probe on frozen encoder vectors moves English {BGE}-small only from $0.588$ to $0.607$, and {BGE}-large-zh from $0.701$ to $0.710$: the information is not ``there but rotated''.

Not every cross-encoder computes identity.
{BGE}-reranker-large ($0.940$), v2-m3 ($0.902$), and base ($0.888$) align with the joint-probe band of \S\ref{sec:comp}.
Jina reranker-v2 ($0.639$) and {MS-MARCO} MiniLM rerankers ($0.545$--$0.558$) also read the pair jointly but stay in the dense band---they score passage relevance, not {PAWS}-style identity.
The gap is therefore not ``cross-encoder beats bi-encoder''; it is that only some joint computations implement the identity operator.

Independently encoded {LM} last-token states tell the same story (Table~\ref{tab:joint}, columns Cos and Sep; cf.\ Figure~\ref{fig:bars}).
Across Llama~3~8{B}, Mistral~7{B}, {Qwen}2.5 (0.5{B}--14{B}), {GPT-2}, {GPT-Neo}-125{M}, and Pythia, English separate cosine at the joint-peak layer is $0.52$--$0.61$, and the best pair probe on those two vectors is $0.51$--$0.62$.
Late fusion, the upper bound of ``just having two vectors'', remains $0.47$--$0.57$ for every modern 3--14{B} model we ran ({Qwen}2.5-14{B} $0.541$): P2 fails, and the 14{B} cosine of $0.614$, the highest geometric number in the table, is still four tenths of {AUC} below the joint probe.

Identity might instead live in the \emph{token sequences}: on the same confirm items, ColBERT-style MaxSim of last-layer tokens reaches $0.62$--$0.71$ across Llama~3~8{B}, Mistral~7{B}, Phi-2, {GPT-J}-6{B}, {OPT}-6.7{B}, and {Qwen}2.5 3{B}--32{B} (best: {Qwen}7{B} at $0.708$, 95\% CI $[0.649,0.767]$); mean-pool cosine sits in $0.57$--$0.64$, and a two-layer {MLP} on the concatenated last-tokens, fit on the $1.2$k discovery pairs, stays at chance ($0.47$--$0.56$).
Scale does not move it into a fixed similarity: 14{B} MaxSim ($0.628$) is \emph{lower} than 7{B}'s, 32{B} MaxSim $0.694$ is $0.24$ below its joint $0.938$, and Phi-2's MaxSim equals 7{B}'s ($0.701$ vs joint $0.893$): MaxSim saturates early as residual lexical matching.

A stronger independent-encoding reader is a small transformer with bidirectional cross-attention over the two \emph{frozen} token sequences---capacity, nonlinearity, and token-level interaction, but no joint {LM} pass; it fits a relatedness control (true vs shuffled partner) at confirm $0.98$--$0.99$ on {Qwen}2.5-7{B} encodings, so it can read the sequences.
Trained for identity on the leak-filtered {PAWS-X} train split it stays at confirm $0.46$--$0.58$ through $20$k pairs on every model; the full $49$k pairs lift it to $0.68$--$0.70$ (Llama~3~8{B}, Mistral~7{B}, {Qwen}2.5 3{B}/7{B}) and $0.87$ ({Qwen}14{B}).
A logistic late fusion of the two sentence vectors stays at $0.52$--$0.58$ on the same $49$k pairs; an {MLP} on the pooled pair reaches $0.72$--$0.75$.
The token sequences carry the material; unlocking it takes a pair computation and forty times the labels the {LM}'s own joint pass needs ($0.90$--$0.95$ from $1.2$k).

Fine-tuning the encoder itself is a different question: after leak-filtering {PAWS-X} train (near-duplicates and generation-template overlap with eval dropped; hyperparameters locked on overlap-matched validation), MiniLM, {BGE}-small, and {BGE}-base reach overlap-matched confirm $0.871$, $0.884$, and $0.930$.
Zero-shot transfer of those checkpoints to overlap-matched {QQP} \emph{falls} relative to the untuned encoders ($0.813{\to}0.701$, $0.824{\to}0.739$, $0.829{\to}0.693$), and {STS-B} Spearman falls from $0.87$--$0.90$ to $0.61$--$0.62$.
A MiniLM \emph{cross}-encoder on the same slices hits {PAWS} confirm $0.906$ only at full train size (chance at $500$ pairs, $0.67$ at $8$k) and still transfers to overlap-matched {QQP} at $0.651$.
Vectors can be trained to carry a {PAWS} identity criterion; they do not, in this protocol, acquire a general meaning-identity geometry, and the fit taxes aboutness.

\section{Identity appears when the pair is computed}
\label{sec:comp}

Table~\ref{tab:joint} (and Figure~\ref{fig:bars}) reports, for each model, the depth of peak English joint confirm {AUC}, together with late fusion, separate cosine, and shuffled joint at that same depth.

\begin{figure}[t]
\centering
\includegraphics[width=0.8\linewidth]{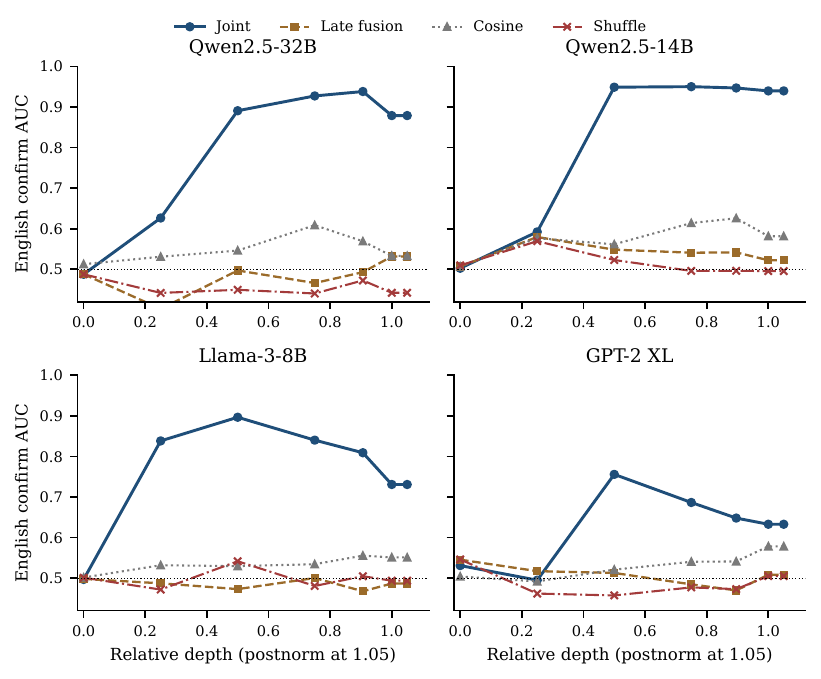}
\caption{Depth of the four readouts.
Joint rises at mid-depth; the other three stay near chance.
{Qwen}14{B} keeps the signal into postnorm ($0.950\to 0.940$); Llama spends it ($0.897\to 0.731$).}
\label{fig:depth}
\end{figure}

\paragraph{Joint is not late fusion, and shuffle kills it.}
For Llama, Mistral, and {Qwen}~$\ge 1.5${B}, joint exceeds late fusion by $0.37$--$0.46$ {AUC} ({Qwen}2.5-14{B}: $0.950$ versus $0.541$); late fusion never leaves the encoder band, so P1 holds and P2 does not.
At the same layer, shuffle {AUC} is $0.46$--$0.54$: the probe reads the pairing, which shuffle destroys, so P3 holds.

\paragraph{Not a decoder-only family effect.}
Table~\ref{tab:joint} is dominated by Llama-style causal models.
We therefore re-ran the same four readouts on the same overlap-matched English bank ($n{=}200$ confirm) on other causal stacks, bidirectional encoders, and encoder--decoders.
Every loaded model satisfies P1 and P3: joint exceeds late fusion, and partner shuffle returns near chance.
Non-causal peaks sit in the main-table band---{Flan}-{T5}-base $0.953$, nli-{DeBERTa} $0.940$, {DeBERTa}-v3 $0.921$---with smaller {T5}/{BART} lower but same-signed ($0.64$--$0.66$).
Nonce substitution of content words preserves the gap.

\paragraph{The wrapper is not the result.}
Four concatenation templates---\texttt{A:}/\texttt{B:}, ``Sentence 1/2'', a raw separator, a blank line---yield joint confirm {AUC} $0.919$--$0.950$ on {Qwen}2.5-7{B} (shuffle $0.47$--$0.52$), $0.933$--$0.961$ on 14{B}, and $0.924$--$0.967$ on 32{B}.
Mistral~7{B} and Llama~3~8{B} at an off-peak layer on a smaller bank are more wrapper-sensitive ($0.776$--$0.881$ and $0.716$--$0.863$), but every wrapper still beats late fusion and collapses under partner shuffle.

\paragraph{QQP is not a {PAWS} artefact.}
On overlap-matched {QQP} (English confirm), joint {AUC} is $0.88$--$0.91$ for every ${\ge}3${B} model we ran ({Qwen}2.5 3{B}--32{B}, Llama, Mistral, Phi-2; {Qwen}7{B} $0.910$, 95\% CI $[0.877,0.943]$), while late fusion rises to $0.72$--$0.77$: duplicate questions share \emph{aboutness}, which {PAWS}'s word-scramble matching forbids and separate vectors can see.
To remove every single-sentence cue by construction we also built an \emph{anchor-balanced} {QQP} bank: each anchor question appears exactly twice, once with a duplicate and once with an overlap-matched non-duplicate, so the first sentence carries no label information and lexical overlap alone scores $0.52$.
There, with the layer locked on development, joint {AUC} is $0.80$--$0.83$ for all seven models ({Qwen}2.5 3{B}/7{B}/14{B}/32{B}, Llama~3~8{B}, Mistral~7{B}, Phi-2; 95\% CIs $\approx\pm0.05$), late fusion $0.55$--$0.60$, separate cosine $0.58$--$0.62$, and partner shuffle $0.42$--$0.46$ ($n{=}300$ confirm).
Geometry can notice that two questions are about the same topic; certifying that they are the same question still takes the joint pass.

\paragraph{The computed direction transfers; the fitted geometry does not.}
Train the linear probe on {PAWS} joint states, lock layer and regulariser on {PAWS} dev, then apply the frozen probe zero-shot: on overlap-matched {QQP} confirm it scores $0.72$--$0.75$ ({Qwen}7{B} / Llama / Mistral / {Qwen}14{B}); on \emph{Chinese} {PAWS} confirm---different language, different wrapper---it scores $0.745$--$0.818$, nearly the in-domain Chinese joint numbers ($0.766$--$0.851$); partner shuffle collapses it ($0.48$--$0.53$).
The reverse direction is sharper: a probe trained only on {QQP} reads {PAWS} confirm at $0.834$ / $0.848$ / $0.724$ / $0.696$ ({Qwen}7{B} / {Qwen}14{B} / Llama / Mistral) without ever seeing a {PAWS} pair---the {Qwen} directions above every independent-encoding readout we could construct on {PAWS} itself (best MaxSim $0.708$), Llama and Mistral at its level.
One linear direction in mid-depth joint states carries the relation across datasets and across languages; the {PAWS}-tuned bi-encoder, by contrast, \emph{lost} {QQP} accuracy relative to its own untuned checkpoint.
Transfer is partial ($0.72$--$0.75$ vs $0.88$--$0.91$ in-domain): part of any probe is dataset-specific.

\paragraph{The relation is a distillable operator; the vectors cannot imitate it.}
If identity is a computed relation, it should behave like an operator: teachable to a smaller joint reader, and unteachable to any function of the two independent vectors---\emph{even with the teacher's own answers as supervision}.
We treat the mid-depth joint direction of a frozen teacher ({Qwen}7{B} / {Qwen}14{B} / Mistral~7{B} / Llama~3~8{B}: {PAWS} confirm $0.934$ / $0.945$ / $0.898$ / $0.895$, zero-shot {QQP} $0.736$ / $0.710$ / $0.812$ / $0.737$) as a soft labeller.
Ridge-fitting the teacher's scores from the \emph{same model's} independently encoded last-tokens (cosine, concatenation, or product/difference features) yields confirm {AUC} $0.46$--$0.62$ and correlation with the teacher between $-0.04$ and $0.20$: the points cannot reproduce the relation the same network computes over them.
A {Qwen}2.5-1.5{B} \emph{joint} student fitted to the same soft scores---no gold labels---reaches {PAWS} confirm $0.85$--$0.90$ under every teacher, the same band as a student trained on gold labels ($0.86$--$0.90$); a 3{B} student from the 7{B} teacher reaches $0.938$ (teacher $0.949$)---the operator is already cheap, and teacher scores are as good as labels.
Distilling on {PAWS} pairs only, the student's zero-shot {QQP} lags the teacher ($0.59$--$0.71$); adding teacher-scored \emph{unlabelled} {QQP} pairs to the distillation set (still no gold) closes most of the gap ($0.71{\to}0.79$ under Mistral, teacher $0.81$; $0.69{\to}0.73$ under Llama; $0.63{\to}0.67$ under 7{B}).
The operator distils from unlabelled pairs while the aboutness encoder stays frozen; installing it in the vectors is the {STS}-taxed fit above.

\paragraph{The circuit is mid-depth.}
Peak joint is not the final residual (Figure~\ref{fig:depth}).
Llama and Mistral peak at layer 16 of 32; {Qwen}7{B} at layer 21 of 28; {Qwen}14{B} at layer 36 of 48; {GPT-2} {XL} at layer 24 of 48.
Postnorm joint is systematically \emph{lower} than the mid-depth peak, but families differ in how much they overwrite: Llama $0.73$ vs $0.90$ ({Qwen}14{B} only $0.940$ vs $0.950$)---{Qwen} keeps what it computed, Llama spends it on next-token prediction.
Either way, the object used as a sentence embedding---the final residual of a separately encoded string---never held the relation, and a final-norm edit that wrecks next-token prediction does not create it: permuting the {RMSNorm} gain on {Qwen}2.5-7{B} yields {KL} $36$ against the untouched head (top-1 agreement $0.4\%$) while separate cosine stays $0.558$ versus $0.561$; Llama, Mistral, 14{B}, and the mute small models all stay in $0.55$--$0.62$ under the same edits.

Chinese confirm repeats the pattern at lower amplitude for multilingual models (joint $0.77$--$0.85$ vs late fusion $0.54$--$0.60$) and vanishes for {GPT-2}, which was not trained on Chinese.

\section{One operator across independently trained models}
\label{sec:universal}

If identity is a computed relation rather than a criterion fitted to a dataset, then models trained by different organisations on different corpora should compute the \emph{same} relation.
Each of six models ({Qwen}2.5 1.5{B}/3{B}/7{B}/14{B}, Llama~3~8{B}, Mistral~7{B}) gets its joint direction located \emph{once}, on {PAWS} English discovery, then frozen; all six score the same held-out items.

\paragraph{The scores agree beyond the labels.}
On {PAWS} confirm, pairwise Spearman between models' item scores averages $0.855$; restricted to cross-family pairs ({Qwen} vs Llama vs Mistral) it is $0.834$.
Agreement is not explained by the binary label: ranking \emph{within} the paraphrases alone, or within the non-paraphrases alone, cross-family agreement is still $0.61$--$0.62$---six models agree on \emph{how much the same} each pair is.

\paragraph{Errors co-occur; consensus adds nothing.}
Where one model errs, the others err: $P(\text{B wrong}\mid\text{A wrong})$ exceeds the base rate by $2.9$--$5.7\times$ on {PAWS}, and the seven items that all six models get wrong are three orders of magnitude more frequent than independent errors would allow.
Averaging the six z-scored outputs does not beat the best single model ($0.963$ vs $0.970$ on {PAWS}; $0.773$ vs $0.778$ on {QQP}): the models share the residual as well as the signal.

\paragraph{The relation is symmetric and graded.}
Swapping the sentences barely moves the score ($\mathrm{score}(A,B)$ vs $\mathrm{score}(B,A)$ Spearman $0.89$--$0.93$ on {PAWS}), and the forward--swap discrepancy carries no label information ({AUC} $0.50$--$0.61$).
Applied zero-shot to {SNLI}, every one of the six directions orders {PAWS} paraphrase $>$ entailment $>$ neutral $>$ contradiction $>$ {PAWS} non-paraphrase, monotonically in the mean, with entailment-vs-contradiction {AUC} $0.79$--$0.98$: a direction located with binary labels measures \emph{degree}.

\paragraph{Unanimous disagreement with {QQP} gold.}
On overlap-matched {QQP} confirm, all six models contradict the published label on $18$ items ($57\times$ the independence rate).
Six of them ($2.4\%$ of the bank) are confident ($|z|>1$).
Consensus {AUC} against those published labels is $0.773$.
Transfer of the frozen {PAWS} coordinate is partial on overlap-matched {MRPC} ($0.48$--$0.61$), and where it transfers less the six models also agree less (frozen-coordinate {QQP}: pairwise Spearman $0.34$--$0.84$, error lift $1.3$--$2.3\times$): the invariant is the relation, and each dataset's coordinate onto it is only partly shared.

\paragraph{The invariant is a sufficient teacher.}
Finally we train a model on the invariant itself: a {Qwen}2.5-1.5{B} reader (four mid-depth blocks unfrozen, linear head on the mid layer) regresses the \emph{consensus} of three cross-family teachers ({Qwen}7{B}, Llama~3~8{B}, Mistral~7{B}; pairwise pool correlation $0.73$--$0.82$) on ${\sim}10$k unlabelled English pairs.
No gold pair label enters anywhere past locating each teacher's coordinate.
The student reaches {PAWS} confirm $0.959$---matching the 7{B} teacher's own probe ($0.956$) and above every single-teacher distillation of \S\ref{sec:comp}---{QQP} $0.774$, and \emph{Chinese} {PAWS} $0.758$ zero-shot from English-only training text: what survives distillation across families and languages is the computed quantity.

\section{Scale, family, and a fossil that grows}
\label{sec:scale}

Two scale stories are easy to confuse: computation as a {Qwen} idiosyncrasy, versus computation as the default organisation of modern {LMs}, which older English {LMs} grow more slowly; Llama~3 and Mistral---not {Qwen}, not instruction-tuned---close the first.
Instruction tuning is not the cause either: {Qwen}2.5-1.5{B} base already reaches $0.916$ (instruct $0.947$; 7{B}-Instruct $0.961$ with late fusion still $0.529$).

Inside {Qwen}2.5 the joint circuit is a scale law with an early ceiling, not a 7{B} accident:
$0.789$ (0.5{B}) $\to$ $0.916$ (1.5{B}) $\to$ $0.937$ (3{B}) $\to$ $0.941$ (7{B}) $\to$ $0.950$ (14{B}) $\to$ $0.938$ (32{B}, 95\% CI $[0.897,0.969]$), with late fusion at 32{B} still chance ($0.493$).
Once the model is large enough to compute the pair, more parameters compute it more cleanly, then stop gaining, and never move the relation into the vectors.

The older families run the same curve more slowly (Table~\ref{tab:joint}, lower block): {GPT-2} climbs $0.665 / 0.688 / 0.652 / 0.756$ (small/medium/large/{XL}) with late fusion pinned near chance---a long plateau, then a jump at {XL}.
The plateau-and-jump repeats in every lineage we ran: TinyLlama-1.1{B} and Phi-1.5 already split joint $0.778$ vs late ${\approx}0.54$ and Phi-2 is that family's jump ($0.893$); SmolLM2 goes $0.753 \to 0.843$; {GPT-Neo} $0.680 \to 0.815$ and {GPT-J}-6{B} $0.868$; {OPT} sits flat at $0.658$--$0.674$ through 1.3{B} and jumps to $0.829$ at 6.7{B}; Pythia stalls at $0.611$--$0.727$; none puts the relation in the vectors.

\section{The mouth is optional}
\label{sec:gen}

A joint probe reads a state.
A user of a chatbot reads tokens.
On a smaller overlap-matched draw ($n{=}150$--$300$ confirm), the two channels come apart by family.

{Qwen} base models can \emph{say} the answer: zero-shot verdict {AUC} $0.915$ at 3{B}, $0.938$ at 7{B} (matching its joint probe $0.941$), $0.960$ at 14{B}, $0.987$ at 32{B}---while length-normalised {PMI} stays $0.47$--$0.61$ throughout (7{B}-Instruct verdict $0.915$: instruction tuning is not what opens the mouth).

Llama~3~8{B} and Mistral~7{B} are the other half of the split: joint probes $0.897$ and $0.915$, verdicts $0.614$ and $0.606$---they compute identity and then refuse to utter it.
Every older or smaller family we ran does the same ({GPT-2} verdicts $0.47$--$0.51$ against joint up to $0.756$; Phi $0.56$--$0.61$ against $0.778$--$0.893$).
Prompting is not a substitute for the joint state: {PMI} does not recover it, and conditional surprisal sits at $0.55$--$0.76$ for every model, unrelated to the joint probe ({OPT}-350{M} $0.76$ against joint $0.66$).

\section{What this is worth in the world}
\label{sec:world}

\paragraph{An end-to-end test where the stakes are visible.}
Dense retrieval compares sentence vectors because it is cheap; on wording twins that comparison is the wrong object.
We built two fully automatic FAQ stacks (frozen {BGE} index, top-10 recall, then a certification step; abstention thresholds set on a disjoint dev split; no human scoring anywhere).
The \emph{Quora bank} (800 canonical {QQP} questions, 425 same-topic hard negatives, 1{,}500 distractors; queries are held-out duplicate partners) is the world where restated questions still share words, so surface overlap is \emph{legitimate} evidence: recall@10 $0.994$, cosine top-1 $0.862$, in/out abstention {AUC} $0.985$.
There the {PAWS}-located direction used \emph{alone} is worse than cosine (top-1 $0.427$): a coordinate located on word-scramble pairs does not rank restated questions, whose twins differ in topic granularity rather than word order; located on 500 in-domain pairs it certifies at $0.800$, and fused with cosine every operator matches cosine's top-1 while lowering false answers on out-of-{KB} queries ($0.125 \to 0.094$): the operator certifies a candidate list, it does not replace recall.
The \emph{trap bank} is the same stack over {PAWS} sentences ({KB} of 2{,}222 entries): 375 of 1{,}289 canonical entries have a word-swap twin \emph{inside} the {KB}, and each of 300 out-of-{KB} queries has its meaning-changed twin \emph{in} the {KB}.
Cosine collapses exactly there: top-1 $0.468$ on trapped queries, and it confidently answers $99\%$ of out-of-{KB} queries---always with the wrong twin.
The frozen {PAWS} direction of \S\ref{sec:comp}, with zero in-domain labels, certifies the same top-10 at $0.83$ on trapped queries; fused with cosine it reaches $0.90$ overall, matching a prompted 7{B} verdict ($0.903$)---a channel only the {Qwen} family can speak (\S\ref{sec:gen})---while confident wrong answers fall from $0.284$ to $0.096$.
The consensus-trained 1.5{B} reader of \S\ref{sec:universal} reproduces this end to end ($0.844$ alone, $0.900$ fused) with no gold labels and no 7{B} at inference.

\begin{table}[t]
\caption{Trap-bank {FAQ} simulation. {KB} of $2{,}222$ entries ($1{,}289$ canonical, $418$ meaning-changed twins, $515$ distractors); $1{,}289$ in-{KB} queries ($375$ with a twin in the {KB}) and $300$ out-of-{KB} queries whose twin is in the {KB}; a random $30\%$ of queries sets thresholds, the rest are reported. All systems see the same cosine top-10. Conf.\ wrong $=$ confidently answered with the wrong entry; false out $=$ answered an out-of-{KB} query at all.}
\label{tab:faq}
\begin{center}
\small
\begin{tabular}{lcccc}
\toprule
System & Top-1 & Top-1 (trap) & Conf.\ wrong & False out \\
\midrule
cosine & $.714$ & $.468$ & $.284$ & $.990$ \\
cosine$\to$7{B} joint probe & $.840$ & $.831$ & $.100$ & $.453$ \\
cosine$+$7{B} joint probe & $.898$ & $.896$ & $.096$ & $.552$ \\
cosine$\to$7{B} prompted verdict & $.903$ & $.903$ & $.088$ & $.547$ \\
cosine$\to$1.5{B} consensus reader & $.844$ & $.842$ & $.109$ & $.502$ \\
cosine$+$1.5{B} consensus reader & $.900$ & $.878$ & $.098$ & $.606$ \\
\bottomrule
\end{tabular}
\end{center}
\end{table}

The practical reading: retrieve with embeddings; certify identity with a joint pass.
The operator is cheap to own, and the aboutness index stays frozen ({BGE}-base {STS-B} Spearman stays $0.895$).

\section{Limitations}

Scope is {PAWS}-style paraphrase under overlap matching.
Independent-encoding readers trail the joint probe even at $49$k training pairs; cross-architecture amplitude varies while the joint-versus-late sign does not; Llama~3 / Mistral compute the relation ($0.90$) without verbalising it ($0.61$).

\section{Conclusion}

Meaning identity on overlap-matched {PAWS-X} is a joint-pass computation, not a property of frozen independent embeddings: the gap holds across causal, encoder, and encoder--decoder families, survives nonce substitution, is shared across training houses, and distils into a cheap reader that repairs a cosine {FAQ} stack.

Code, scripts, and result dumps will be released in a public repository in a subsequent update of this preprint.

\input{meaning_identity.bbl}
\end{document}

%% file: math_commands.tex
\def\eqref#1{equation~\ref{#1}}

\def\1{\bm{1}}

\DeclareMathAlphabet{\mathsfit}{\encodingdefault}{\sfdefault}{m}{sl}
\SetMathAlphabet{\mathsfit}{bold}{\encodingdefault}{\sfdefault}{bx}{n}

